\documentclass[runningheads,a4paper]{llncs}

\usepackage{amssymb}
\usepackage{graphicx}
\usepackage{enumerate}
\usepackage{paralist}
\usepackage{hyperref}
\usepackage{url}
\usepackage{tabularx}
\usepackage[utf8]{inputenc}
\usepackage{svg}
\usepackage{subcaption}

\usepackage[ruled,section]{algorithm}
\usepackage[noend]{algorithmic}
\usepackage{amssymb}
\usepackage{fca}
\usepackage{amsmath}

\usepackage{amsfonts}
\usepackage{alltt}

\usepackage{url}
\usepackage{multirow}
\urldef{\mailsa}\path|dalyutkin@gmail.com|
\urldef{\mailsb}\path|dvpozdnyakov@hse.ru|
\urldef{\mailsc}\path|mumalik@hse.ru|
\urldef{\mailsd}\path|dignatov@hse.ru|
\newcommand{\keywords}[1]{\par\addvspace\baselineskip
\noindent\keywordname\enspace\ignorespaces#1}

\graphicspath{{arxiv-pic-prez/}}
\svgpath{{arxiv-pic-prez/}}

\begin{document}

\mainmatter  %
\titlerunning{Transformer-based classification of user queries for medical consultancy with respect to expert specialization}

\title{Transformer-based classification of user queries for medical consultancy with respect to expert specialization}

\titlerunning{Transformer-Based Classification of User Queries for Medical Consultancy}

 \author{Dmitry Lyutkin\inst{1} \and Andrey Soloviev \inst{2} \and Dmitry Zhukov \inst{2} \and Denis Pozdnyakov \inst{1} \and Muhammad Shahid Iqbal Malik\inst{1} \and Dmitry I. Ignatov\inst{1}}
\authorrunning{Dmitry Lyutkin et al.}

\institute{National Research University Higher School of Economics, Moscow\\
\mailsa, \mailsb, \mailsc, \mailsd \\
\and
Babyblog LTD, Moscow
}

\toctitle{Transformer-Based Classification of User Queries for Medical Consultancy}
\tocauthor{}
\maketitle

\begin{abstract}
The need for skilled medical support is growing in the era of digital healthcare. This research presents an innovative strategy, utilizing the RuBERT model, for categorizing user inquiries in the field of medical consultation with a focus on expert specialization. By harnessing the capabilities of transformers, we fine-tuned the pre-trained RuBERT model on a varied dataset, which facilitates precise correspondence between queries and particular medical specialisms. Using a comprehensive dataset, we have demonstrated our approach's superior performance with an F1-score of over 92\%, calculated through both cross-validation and the traditional split of test and train datasets. Our approach has shown excellent generalization across medical domains such as cardiology, neurology and dermatology. This methodology provides practical benefits by directing users to appropriate specialists for prompt and targeted medical advice. It also enhances healthcare system efficiency, reduces practitioner burden, and improves patient care quality. In summary, our suggested strategy facilitates the attainment of specific medical knowledge, offering prompt and precise advice within the digital healthcare field.

\end{abstract}

\section{Introduction}

The demand for qualified medical assistance has never been more significant, especially in the digital era. As online platforms increasingly serve as crucial sources of medical information and support \cite{song2016trusting}, ensuring the provision of accurate and specialized advice becomes imperative. One such platform that has garnered attention is Babyblog.ru \cite{babyblog}, which uniquely leverages user-generated content as a gateway and contextual backdrop for medical professionals' knowledge dissemination.

However, the abundance of user-generated content poses challenges regarding the scientific credibility and reliability of the information shared \cite{keshavarz2021evaluating}. Consequently, there is a pressing need to implement mechanisms that ensure the verification and enrichment of user-generated content through the input and recommendations of diverse professionals, including doctors, psychologists, speech therapists, and educators. This collaborative approach allows professionals to review user posts, comments, and discussions, thereby providing expert insights, correcting non-specialist advice, and ensuring the delivery of accurate and reliable medical information.

Given the substantial volume of user-generated content across various platforms, encompassing a wide array of topics including medical, quasi-medical, and non-medical domains, the challenge of identifying content requiring medical or professional verification becomes increasingly significant. Furthermore, the importance of classifying this diverse content based on thematic specialization emerges as a critical factor in directing relevant user queries to the appropriate professionals for verification purposes.

To address these challenges, the research and development team embarked on the development of an automatic classifier for medical texts. This classifier aims to determine the likelihood of associating a given text with a specific medical specialization. The envisioned implementation involves integrating the classifier into the platform, wherein it identifies medical content and assigns corresponding medical specializations. Subsequently, professionals in the respective specializations are notified to verify the content and provide appropriate responses.

The successful development of the classification system offers multiple benefits, including streamlining the verification process by reducing irrelevant information presented to medical professionals, alleviating the workload involved in content verification, and accelerating the provision of professional responses to users. Moreover, the proposed system serves as a valuable tool in improving the quality, completeness, and reliability of medical information related to conception, pregnancy, and motherhood on the platform.

This study aims to explore the efficacy of a transformer-based system in classifying user-generated medical content within the context of Babyblog.ru. By leveraging advanced Natural Language Processing (NLP) techniques, this research endeavors to revolutionize the ways users access specialized medical expertise, ensuring the delivery of timely and accurate guidance while upholding scientific rigor and reliability.

\section{Related works}

In the realm of medical text classification, the research paper titled "Automatic Medical Specialty Classification Based on Patients’ Description of Their Symptoms" \cite{relatedFirst:2023} presents a significant contribution to the field. The study proposes a pioneering Hybrid Model (HyM) that combines multiple deep learning techniques, including LSTM, TEXT-CNN, BERT, and TF-IDF, along with an attention mechanism to address the critical challenge of accurately directing patients to the appropriate medical specialty based on their symptom descriptions.

The article "Text Classification Using Improved Bidirectional Transformer" \cite{relatedSecond:2022} presents a significant contribution to the field of text processing, particularly in the context of handling large amounts of text data generated daily. The authors highlight the necessity for automation in text data handling and discuss recent developments in text processing, including attention mechanisms and transformers, as promising methods to address this need.

In their study, the authors introduce a novel bidirectional transformer (BiTransformer) model, constructed using two transformer encoder blocks that utilize bidirectional position encoding. By considering both forward and backward position information of the text data, the proposed BiTransformer aims to capture more comprehensive contextual dependencies, enhancing the model's ability to handle complex text data.

To evaluate the effectiveness of attention mechanisms in the classification process, the authors compare four models, namely long short-term memory (LSTM), attention, transformer, and their proposed BiTransformer. Experiments are conducted on a large Turkish text dataset comprising 30 categories, allowing for a comprehensive assessment of the models' performance.

One of the notable findings of the study is the promising results obtained from the classification models that employ transformer and attention mechanisms compared to classical deep learning models. This demonstrates the potential of attention mechanisms and transformers in text classification tasks, showcasing their ability to capture meaningful patterns and context in textual data.

The authors also investigate the impact of using pretrained embeddings on the models' performance. Pretrained embeddings, which capture semantic representations of words based on large corpora, have been a popular approach to improve model performance in various NLP tasks. The study sheds light on how pretrained embeddings can further enhance the efficiency and accuracy of text classification models.

Perhaps the most significant result of the study is the superior performance of their proposed BiTransformer in text classification. By effectively incorporating bidirectional position encoding and leveraging transformer-based architecture, the BiTransformer outperforms other models in accurately categorizing the text data.

"Text Classification Using Improved Bidirectional Transformer" provides insights into the potential of attention mechanisms and transformers in text processing. The introduction of the BiTransformer and its superior performance in text classification open up new avenues for future research and application of transformer-based models in NLP tasks. The study's findings have important implications for automating text data handling, sentiment analysis, information retrieval, and other text-related applications. As the demand for efficient and accurate text processing techniques continues to grow, this research makes a significant contribution to the advancement of the field and serves as a valuable reference for researchers and practitioners in the domain of natural language processing.

\section{Data Collection: Building a Comprehensive Dataset for Medical Text Classification}\label{review}

In this section, we describe the process of building the dataset. It includes developing data parsers and normalizers to create a normalized dataset for the experimental setup. 

\subsection{Data parsing}

To obtain a suitable training sample, we extensively explored various Russian-language websites that provide public access to medical questions posed by users to healthcare professionals. We employed specific criteria to select our data sources, including: 1) presence of openly accessible sections containing medical questions, 2) availability of pre-annotated questions based on medical specialization, and 3) responses provided by healthcare professionals, which verified the appropriateness of the assigned medical specialization to the responding doctor.

Based on our analysis, we selected the following sources for data acquisition: \textbf{sprosivracha.com} \cite{sprosivracha}, \textbf{doctu.ru} \cite{doctu}, \textbf{03online.com} \cite{03online} and \textbf{health.mail.ru} \cite{healthmail}. We developed software for parsing these data sources, allowing for the asynchronous, multi-threaded retrieval of information from public data sources. The software was designed to extract relevant information from the HTML structure and store them for further processing.

\begin{table}[ht]
\centering
\caption{Comparison of Medical Question Platforms}
\begin{tabularx}{\textwidth}{|>{\centering\arraybackslash}p{0.3\textwidth}|>{\centering\arraybackslash}p{0.41\textwidth}|c|}
\hline
\textbf{Website} & \textbf{Number of Questions} & \textbf{Percentage of Total} \\
\hline
sprosivracha.com & 550,000 & 23.2 \\
\hline
doctu.ru & 83,000 & 3.5 \\
\hline
03online.com & 1,148,000 & 48.4 \\
\hline
health.mail.ru & 590,000 & 24.9 \\
\hline
\end{tabularx}
\end{table}

In the subsequent step, the algorithm asynchronously processes each row of the obtained table and retrieves the HTML code of the page containing the question posed to the doctor. From each HTML code, the algorithm extracts the question text and the doctor's specialization using predefined tags and classes that enclose the relevant text.

The extracted data (question text and doctor's specialization) are then added to the same table, complementing the rows with the data source (URL as the data source identifier).

Once the parsing process and table population are complete, all the acquired data are exported to a CSV file for further processing.

\subsection{Data Augmentations}

After analyzing the acquired dataset, we noticed that the distribution of data units across medical specializations followed a pattern similar to a Pareto distribution. This observation can be attributed to the fact that certain medical specializations are in higher demand compared to others, resulting in a significant number of data units belonging to those specific classes. However, to ensure the stability and resilience of our classifier, it was crucial to address the class imbalance issue \cite{johnson2019survey}.

To tackle class imbalance and enhance the model's generalization ability, we employed data augmentation methods facilitated by the Albumentations library \cite{albumentations:2018}. This versatile tool enabled us to create new textual data by rearranging word positions within sentences, preserving the overall context and meaning. This diversification of input data aimed to produce a more balanced and comprehensive dataset. Specifically, our data augmentation techniques involved shuffling words and reordering sentence components.

Through the augmentation process, we were able to generate additional data points for the minority classes, effectively reducing the class imbalance and achieving a more uniform distribution across all medical specializations. This augmentation strategy not only helped to improve the classifier's performance for underrepresented classes but also enhanced its ability to handle unseen data during the testing phase.

After applying all the necessary transformations and augmentations, we successfully obtained a dataset with a more uniform distribution of classes and expanded the original dataset to 5 million texts, where there are approximately 50000 exemplars per class. This balanced dataset formed the basis for training and evaluating our proposed framework.

\begin{figure}[H]
 \centering
 \includegraphics[width=\textwidth]{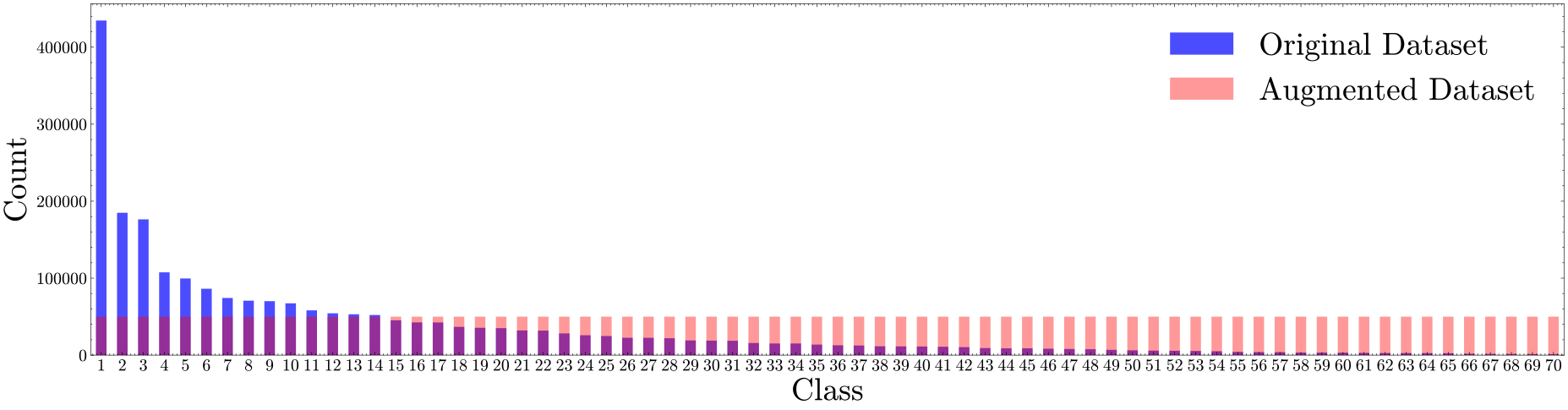}
 \caption{Class Distribution after transformation and augmentation (first 70 classes).}
 \label{fig:class_distrib_after}
\end{figure}

The development of the proposed dataset arises from the recognition of a crucial need in the field. While there exist analogous datasets, they exhibit certain limitations in adequately covering a comprehensive spectrum of diseases and medical conditions. Additionally, these existing datasets suffer from a paucity of records, which impedes their capacity to comprehensively represent the diverse range of health concerns. A further challenge lies in the nature of the content within these datasets; predominantly composed in technical language, they lack congruence with the narratives of individuals detailing their ailments. This discrepancy hampers the efficacy of these datasets in capturing the nuanced descriptions of health issues as articulated by individuals themselves. In light of these deficiencies, the development of the proposed dataset emerges as a pivotal endeavor, with the intent to address these gaps and furnish a resource that aligns more closely with the authentic narratives of people regarding their health conditions. Through the proposed dataset, an avenue is created to elicit novel insights that may have remained obscured within the confines of the existing datasets, fostering a more holistic understanding of individuals' health experiences.

\section{Proposed Methodology}\label{sec:review}

\begin{figure}[H]
 \includegraphics[width=\textwidth]{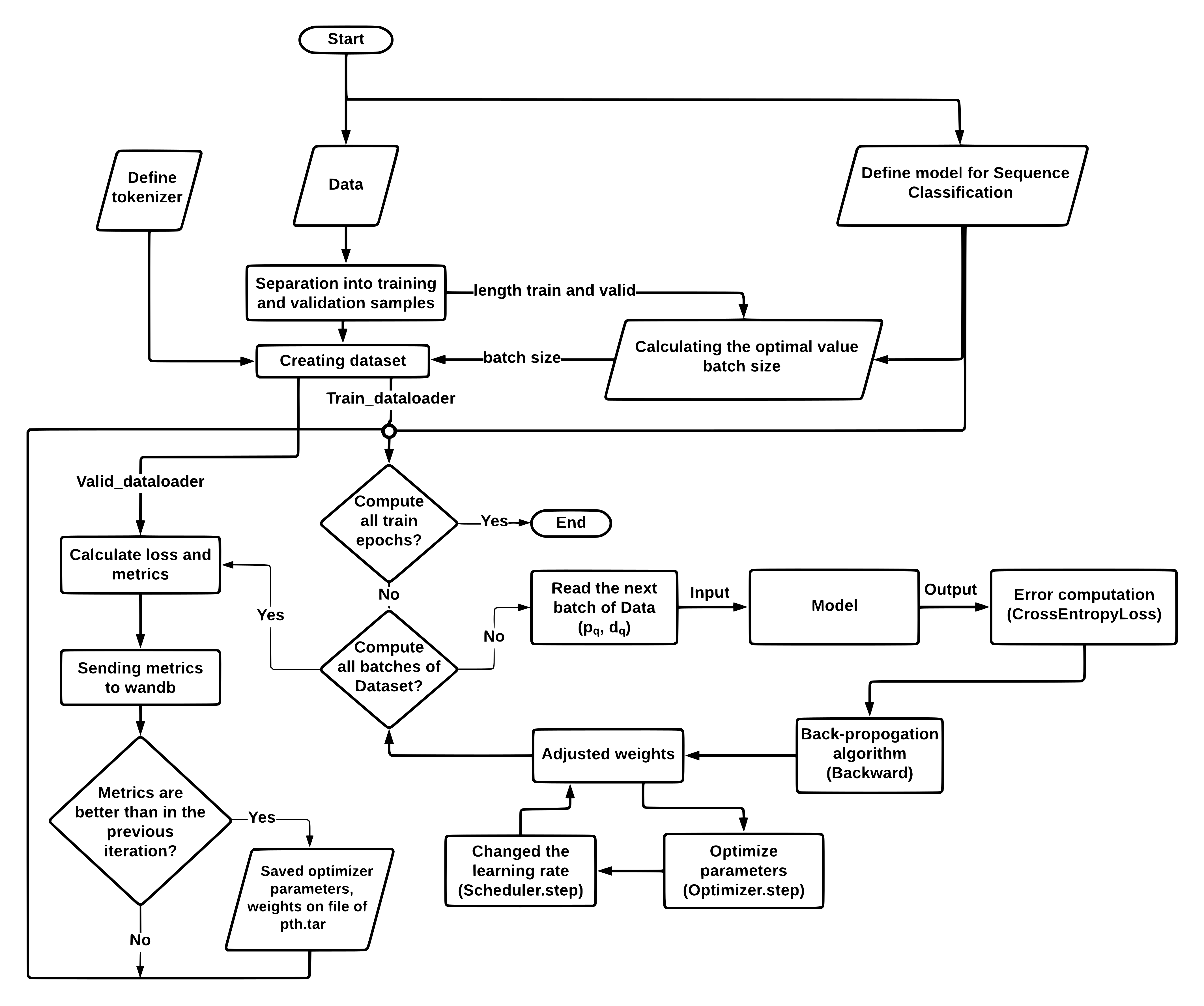}
 \caption{Processing pipeline.}
 \label{fig:pipeline}
 \centering
\end{figure}

This section provides details of proposed methodology. We explore various methods of transformers and their training. The pipeline is presented in Figure \ref{fig:pipeline}.

\subsection{Transformer Models}

Typically, neural networks are trained using the backpropagation algorithm \cite{backprop:1992}, which optimizes model parameters by computing gradients to improve generalization performance through error minimization and/or enhancing metrics on the validation set. However, this method heavily relies on the choice of optimization algorithm \cite{text_classification:2020}, as there is a risk of getting stuck in local minima during gradient computation, leading to the model's inability to learn and improve prediction/recognition quality (vanishing gradients). To address this, we employed the AdamW \cite{adamw:2022} optimizer, one of the state-of-the-art methods, which leverages information about the learning rate history to approximate the direction of the anti gradient while incorporating momentum to expedite the convergence of our function. This optimizer significantly improves model training; however, it is sensitive to the choice of the learning rate. Hence, we employed a learning rate scheduler that suits our task best - the cosine scheduler \cite{cosine:2021}. This scheduler adjusts the learning rate for each batch of data, allowing transformers to adaptively change the learning rate. We opted for the cross-entropy loss function as our choice for the loss function, as it measures how well the model is trained for classification tasks. The utilization of the AdamW optimizer and the linear scheduler with a warm-up for training text classifiers based on BERT has proven effective for several reasons:

\textbf{AdamW Optimizer:} AdamW is a variant of the Adam optimizer that has been shown to work well for fine-tuning pre-trained models such as Transformer \cite{attention:2017}. It addresses the weight decay issue in Adam, helping to prevent overfitting \cite{adamw_adv:2022}.

\textbf{Cosine Scheduler:} The cosine scheduler modifies learning rate starting with a lower learning rate and gradually increases it over a specified number of training steps. This warm-up period allows the model to converge faster, preventing instability or fluctuations in the loss function during training \cite{cosine_adv:2021}.

It is worth noting that the optimal training methods may vary depending on the specific task and the data being used, requiring experimentation and fine-tuning.

Furthermore, there are several reasons why transformer-based models have emerged as the preferred choice for medical text classification compared to classical machine learning methods \cite{li2021survey:2021}:

\textbf{Pretraining:} Transformers are pre-trained on large corpora of texts, which provides them with a strong knowledge base and an understanding of language patterns and word relationships. This pretraining allows models like Transformer to perform well across various NLP tasks with limited fine-tuning.

\textbf{Contextual Representation:} Transformers employ bidirectional attention mechanisms to create contextual word representations, enabling them to capture the context and meaning of words within a sentence. This is particularly crucial for text classification, where understanding sentence context is key to assigning the correct label.

\textbf{Transfer Learning:} The pretraining and fine-tuning process of BERT allows for transfer learning, where a pre-trained model on a related domain can be accurately fine-tuned for specific tasks with a limited amount of labeled data. This is a significant advantage for text classification tasks, which often have a limited number of annotated data.

\textbf{Superior Performance:} Transformers have shown to outperform traditional machine learning methods in various NLP tasks, including text classification. This can be attributed to their ability to capture contextual representations and word relationships, which are crucial for understanding sentence semantics.

\textbf{Pretraining on Russian Texts:} Models pre-trained on large Russian corpora exhibit improved performance and quality in Russian text processing tasks compared to training from scratch. Raw textual data provides models with a natural foundation for building language contextual representations. The size of the raw text corpus is crucial during the pretraining phase.

It is important to note that traditional machine learning methods are still widely used and can yield good results for specific NLP tasks. However, the possibilities offered by pretraining, contextual representation, transfer learning, availability of Russian language models, and the use of raw texts for training make transformers a powerful tool for medical text classification.

\subsection{Training Process}

The training algorithm makes use of the architecture and pre-trained weights of a transformer model, obtained from the transformers \cite{transformers:2020} package, and cached for subsequent utilization. During this phase, the model initialization is executed, which includes the initiation of the tokenizer via the AutoTokenizer module from the transformers library. Additionally, the output layer of the model is modified to suit the specific task at hand.

Subsequently, an optimal batch size is determined by generating an artificial dataset and conducting a grid search to identify the batch size that optimally balances computational efficiency and resource utilization. This strategic step is essential to ensure the model's efficiency during computations on the server.

During the course of training, a significant aspect involves the aggregation of energy following the application of the softmax activation function \cite{maida2016cognitive}. This process offers insights into the model's confidence levels for each distinct class. The resulting energy accumulation, presented in the form of probability scores, functions as an indicator of the model's assurance in assigning input data to specific classes. This measure of confidence holds a central role in the model's final predictions, contributing to its ability to make well-informed decisions about the designated classes. It's important to mention that the target labels are numerical class identifiers, previously encoded using the LabelEncoder, while the input data comprises natural language questions with descriptive explanations of medical issues.

\section{Experimental Setup}

In this section, we describe the detail of experimental setup including the hardware setup used for training the models. Furthermore the training time required for each transformer model is also discussed.

\subsection{Hardware Setup}

For the model training, we utilized a powerful hardware setup consisting of two NVIDIA V100 GPUs with 32GB of memory each. The GPUs were complemented by 250GB of RAM, ensuring efficient processing and storage of the large-scale dataset. The training process was conducted on the high-performance computing system cHARISMa \cite{hpc:2021}, which provided the necessary computational resources for training deep learning models.

\subsection{Training Time}

The training time for each transformer model varies depending on its architecture and complexity. Following are the training times observed for each model:

\begin{itemize}
  \item \textbf{SBERT \cite{sbert:2019} }: The SBERT model required approximately 54 hours to complete the training process. The extensive training time can be attributed to its deep architecture and complex attention mechanisms.  
  \item \textbf{LaBSE \cite{LaBSE:2022}}: The LaBSE model demonstrated faster training times, with the training process taking approximately 12 hours. The model's efficient architecture and advanced pre-training techniques contribute to its reduced training time.
  \item \textbf{RuBERT \cite{rubert:2019} }: Training the RuBERT model took around 13 hours. The model's architecture, specifically designed for the Russian language, required additional time for fine-tuning and convergence.
  \item \textbf{BERT \cite{bert:2019} }: Similar to LaBSE, the BERT model also completed training in approximately 12 hours. Its widely adopted architecture and availability of pre-trained weights contribute to the faster training time.
  \item \textbf{BART \cite{bart:2019}}: The BART model, known for its transformer-based sequence-to-sequence architecture, required a longer training time of 55 hours. The complexity of the model and the additional training required for the encoder-decoder structure contributed to the extended duration.
\end{itemize}

The complete training and evaluation cycle, including cross-validation with a k-fold value of 3, ranged from 3 days to 12 days, depending on the specific model. This timeframe accounted for multiple iterations of training, hyperparameter tuning, and performance evaluation.

The significant training times for certain models underscore the computational resources and time investment required for training large-scale transformer models. However, the improved performance achieved by these models justifies the efforts put into training and fine-tuning them.

In the next section, we present the results of our experiments and evaluate the performance of each model.

\section{Experimental Results and Performance Analysis}
In this section, we present the analysis of learning outcomes obtained by several experiments. We trained several models using cross-validation techniques and evaluated their performance using the F1-score metric.

\begin{figure}[ht]
 \centering
 \includegraphics[width=4.5in]{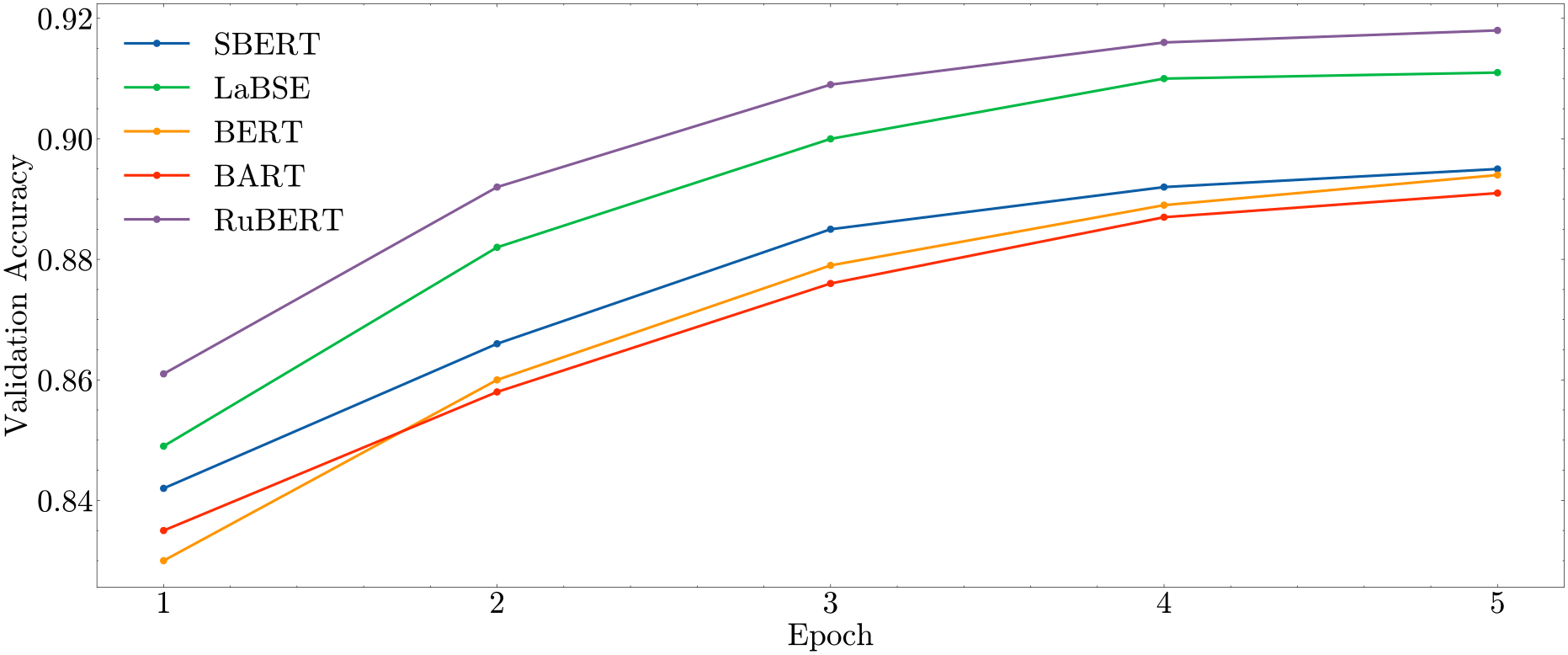}\\
  \caption{Training Curve of Various Models across Folds.}
  \label{fig:training_curve}
\end{figure}

As depicted on Figure \ref{fig:training_curve}, the plot presents the learning curves of the LaBSE, SBERT, BERT, and BART models. It is evident from the graph that LaBSE demonstrates remarkable performance superiority compared to the other models. The learning curve of LaBSE displays significantly higher accuracy and faster convergence, indicating its exceptional capability to learn from the provided dataset. However, for the Russian text specifically, the RuBERT model achieves the highest quality due to its pre-training on a Russian corpus of texts.

Conversely, the learning curves of SBERT, BERT, and BART models exhibit relatively lower accuracy and slower convergence, suggesting their relatively inferior performance in this specific task. The notable contrast in performance between LaBSE and RuBERT underscores their effectiveness and underscores their potential as robust models for the given classification problem.

This can be explained by the fact that LaBSE is good at distinguishing between entities, this can be seen in the Umap image, which converts sentence embeddings into a two-dimensional representation. Also, Umap shows that RuBERT is very similar to other pictures which show rather poor quality, but considering that this model is well adapted for Russian, after fine-tuning it starts to show much better quality.
\begin{figure}[ht]
  \centering
  \begin{subfigure}{0.3\linewidth}
    \centering
    \includegraphics[width=\linewidth]{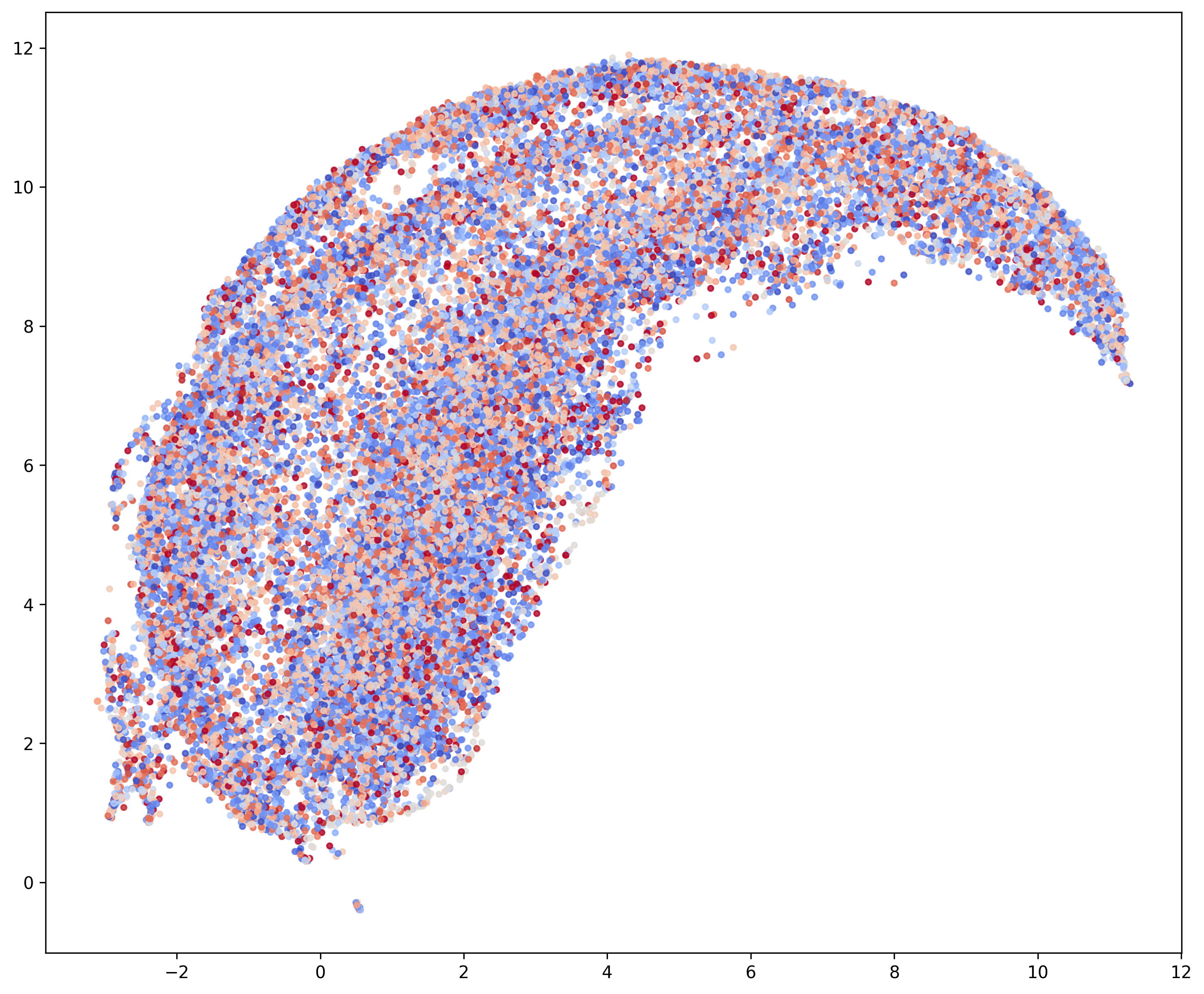}
    \caption{RuBERT}
  \end{subfigure}
  \hfill
  \begin{subfigure}{0.3\linewidth}
    \centering
    \includegraphics[width=\linewidth]{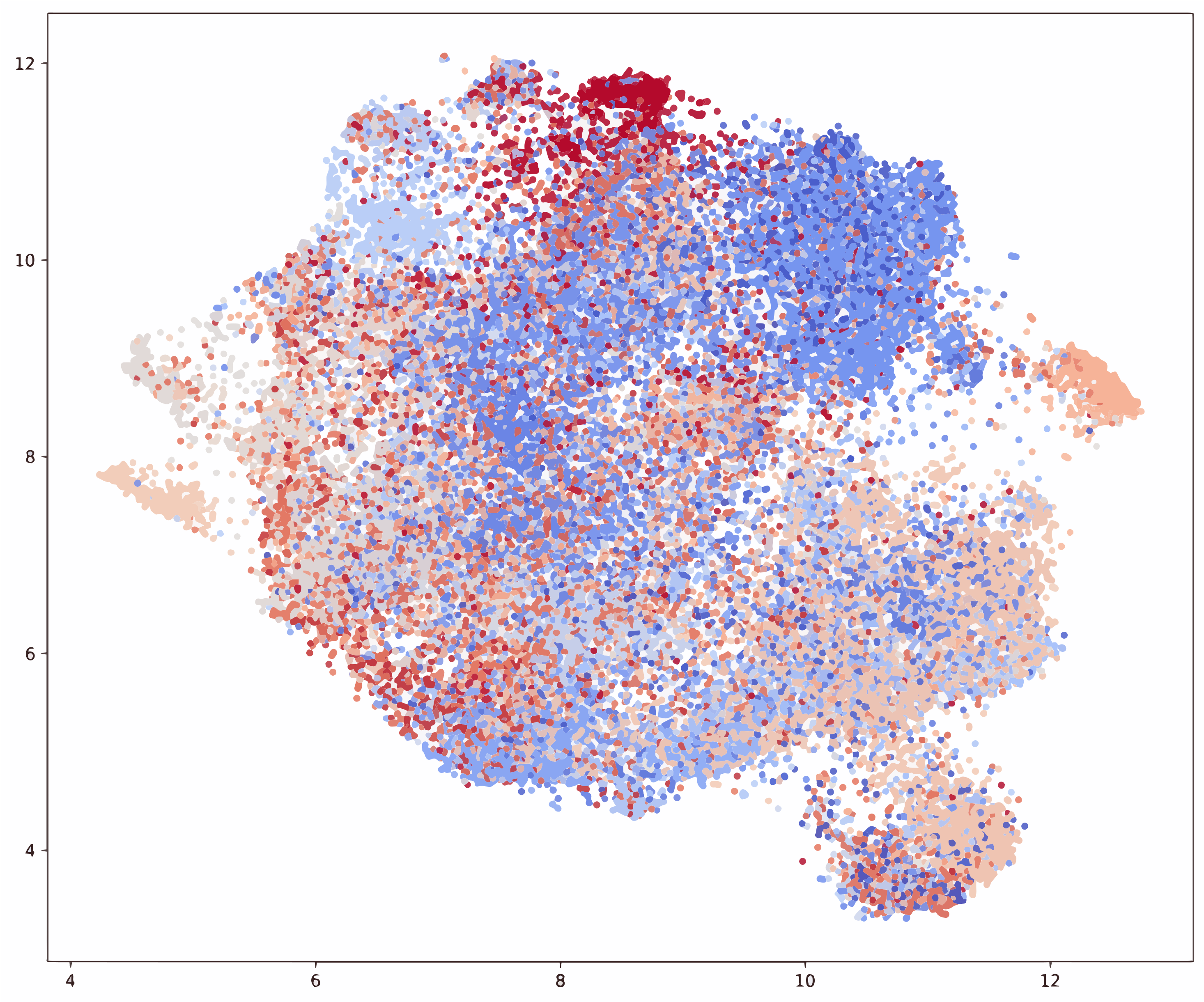}
    \caption{LaBSE}
  \end{subfigure}
  \hfill
  \begin{subfigure}{0.3\linewidth}
    \centering
    \includegraphics[width=\linewidth]{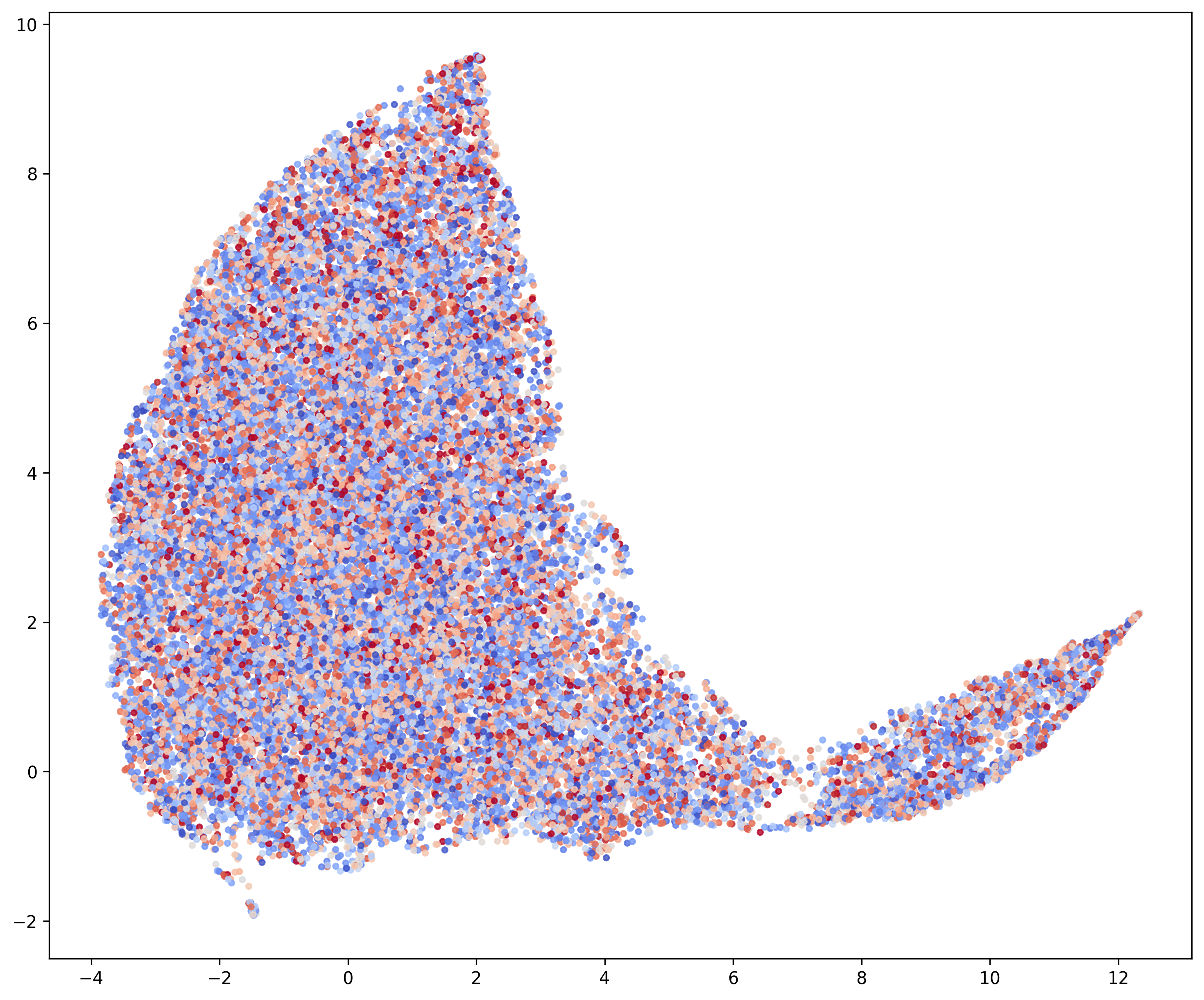}
    \caption{BERT}
  \end{subfigure}
\end{figure}
\begin{figure}[ht]
  \ContinuedFloat
  \centering
  \begin{subfigure}{0.3\linewidth}
    \centering
    \includegraphics[width=\linewidth]{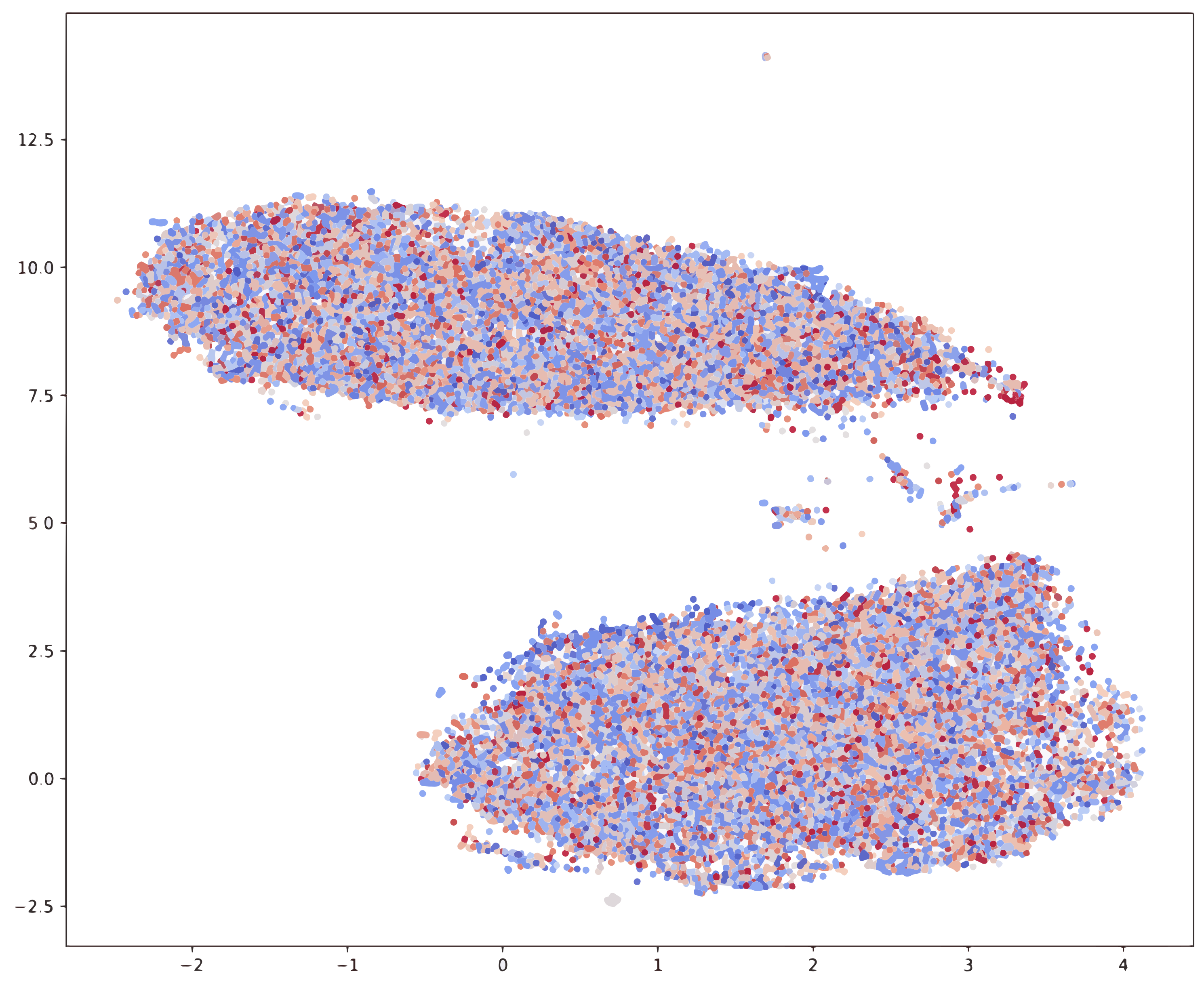}
    \caption{BART}
  \end{subfigure}
  \hfill
  \begin{subfigure}{0.3\linewidth}
    \centering
    \includegraphics[width=\linewidth]{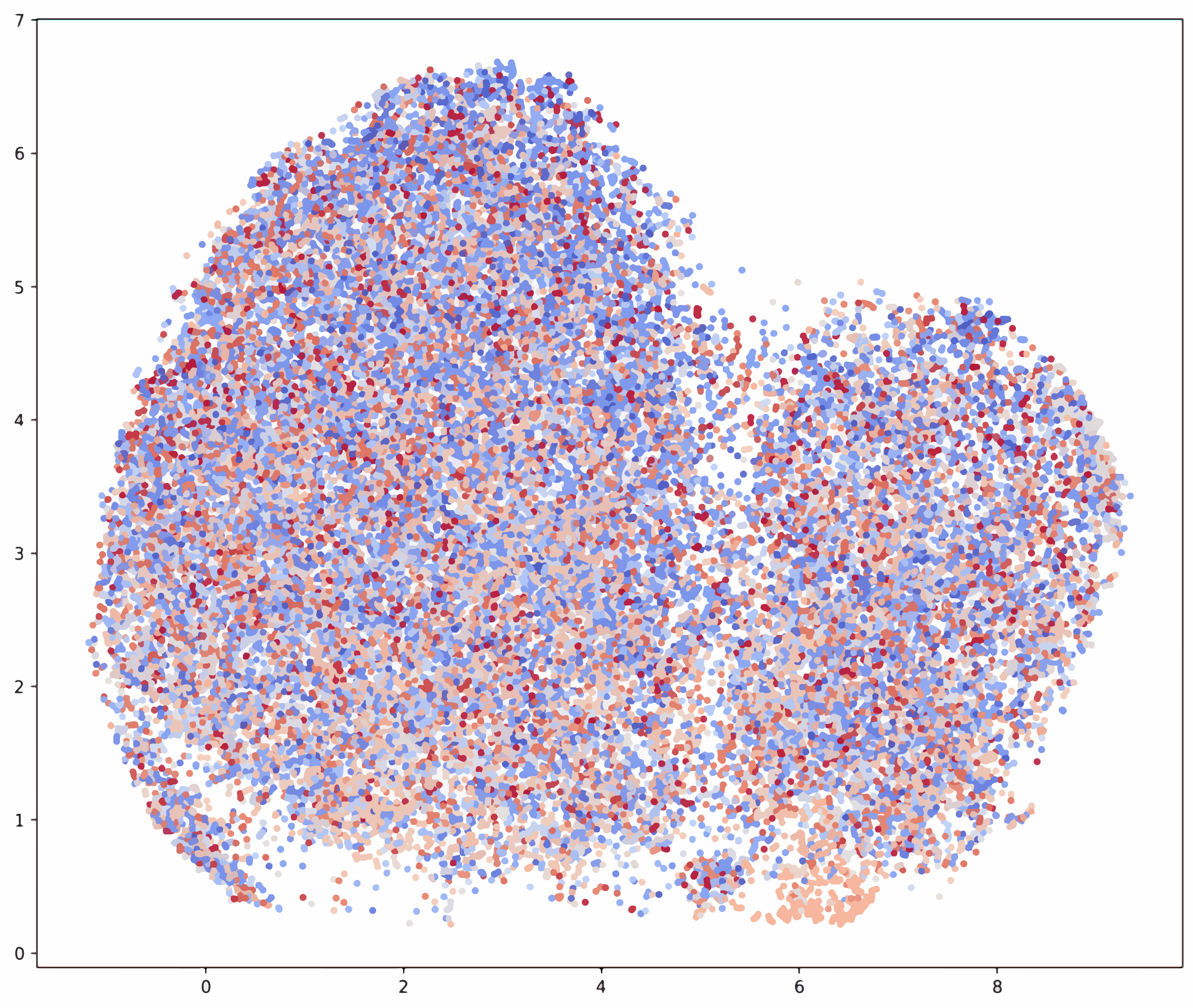}
    \caption{SBERT}
  \end{subfigure}
  \caption{Umap image of different models on our dataset}
\end{figure}
  
\newpage

The following models were trained and their corresponding F1-scores are reported.

\begin{table}[H]
\centering
\caption{Performance Comparison of Transformer Models}
\label{tab:performance_comparison}
\begin{tabularx}{\textwidth}{|>{\centering\arraybackslash}X|>{\centering\arraybackslash}X|>{\centering\arraybackslash}X|>{\centering\arraybackslash}X|>{\centering\arraybackslash}X|}
\hline
\multirow{2}{*}{\centering Model} & \multicolumn{2}{c|}{K-fold (F1-score, k = 3)} & \multicolumn{2}{c|}{Split (F1-score, train = 90\%)} \\
\cline{2-5}
& not augmented & augmented & not augmented & augmented \\
\hline
BART & 0.798 & 0.891 & 0.794 & 0.896 \\
\hline
BERT & 0.743 & 0.894 & 0.760 & 0.903 \\
\hline
LaBSE & 0.824 & 0.911 & 0.833 & 0.913 \\
\hline
LogRegression & 0.457 & 0.552 & 0.531 & 0.564 \\
\hline
Random Forest & 0.521 & 0.579 & 0.596 & 0.603 \\
\hline
\textbf{RuBERT} & \textbf{0.839} & \textbf{0.918} & \textbf{0.852} & \textbf{0.918} \\
\hline
SBERT & 0.782 & 0.905 & 0.761 & 0.895 \\
\hline
SVM & 0.525 & 0.565 & 0.534 & 0.598 \\
\hline
\end{tabularx}
\end{table}

Results in table \ref{tab:performance_comparison} provide insights into the performance of different models in our experimental setup. The high F1-scores obtained by RuBERT and LaBSE suggests that they effectively captured the semantic representations and contextual information in the text data. SBERT and BERT also demonstrated competitive performance, although slightly lower than RuBERT and LaBSE. BART exhibited a slightly lower F1-score, indicating that its performance may be influenced by the specific task and dataset. Overall, the analysis of learning outcomes highlights the effectiveness of various models in our experiments, with RuBERT and LaBSE demonstrating particularly promising results. These findings contribute to our understanding of the strengths and limitations of different models and can guide future research and practical applications in the field of NLP.

\begin{table}[H]
\centering
\caption{Performance Evaluation of RuBERT for Medical Specialties Classification}
\label{tab:performance_evaluation}
\begin{tabular}{l|cccc}
\textbf{Category} & \textbf{Precision} & \textbf{Recall} & \textbf{F1-Score} & \textbf{Support} \\
\hline
ENT & 0.7555 & 0.7432 & 0.7493 & 15276 \\
Ophthalmologist & 0.9403 & 0.9210 & 0.9305 & 14936 \\
Pediatric Surgeon & 0.8405 & 0.8782 & 0.8589 & 14847 \\
Gynecologist & 0.7834 & 0.7459 & 0.7642 & 14844 \\
Dentist & 0.8815 & 0.8893 & 0.8854 & 14861 \\
Sexologist-Andrologist & 0.7904 & 0.6955 & 0.7399 & 15148 \\
Therapist & 0.5066 & 0.3738 & 0.4302 & 15080 \\
Surgeon & 0.6705 & 0.5818 & 0.6230 & 14929 \\
Cardiologist & 0.8646 & 0.8567 & 0.8606 & 14836 \\
Psychologist & 0.7759 & 0.7215 & 0.7477 & 15020 \\
Orthopedic Traumatologist & 0.7981 & 0.7683 & 0.7829 & 15081 \\
Pediatrician & 0.6482 & 0.5712 & 0.6073 & 15087 \\
Dermatologist & 0.7111 & 0.6569 & 0.6829 & 14941 \\
Neurosurgeon & 0.8797 & 0.9025 & 0.8910 & 14898 \\
Endocrinologist & 0.8478 & 0.8072 & 0.8270 & 15011 \\
Venereologist & 0.7763 & 0.8112 & 0.7934 & 15140 \\
Urologist & 0.6445 & 0.6240 & 0.6341 & 15110 \\
Neuropathologist & 0.6633 & 0.5834 & 0.6206 & 15058 \\
Medical Doctor & 0.8667 & 0.8824 & 0.8745 & 14959 \\
Infectious Disease Specialist & 0.8409 & 0.7986 & 0.8192 & 14924 \\
Oncologist & 0.8796 & 0.8742 & 0.8769 & 14957 \\
Gastroenterologist & 0.7574 & 0.7339 & 0.7455 & 14839 \\
. . . \\
\textbf{accuracy} & 0.9111 & 0.9111 & 0.9111 & 0.9031 \\
\textbf{macro avg} & 0.9177 & 0.9205 & 0.9189 & 1470000 \\
\textbf{weighted avg} & 0.9178 & 0.9201 & 0.9189 & 1470000 \\
\end{tabular}
\end{table}

Table \ref{tab:performance_evaluation} presents the performance evaluation results of a classification model for various categories of medical specialists. It includes the metrics precision ($P$), recall ($R$), F1-score ($F1$), and support for each category. These metrics provide insights into the model's ability to correctly classify instances belonging to different medical specialties.

These evaluation metrics help assess the effectiveness of the classification model in distinguishing between different medical specialties. The values in the table represent the performance of the model for each category, allowing for a comparison of its accuracy and effectiveness across various medical specialties.

\begin{figure}[H]
 \centering
\includegraphics[width=4.5in]{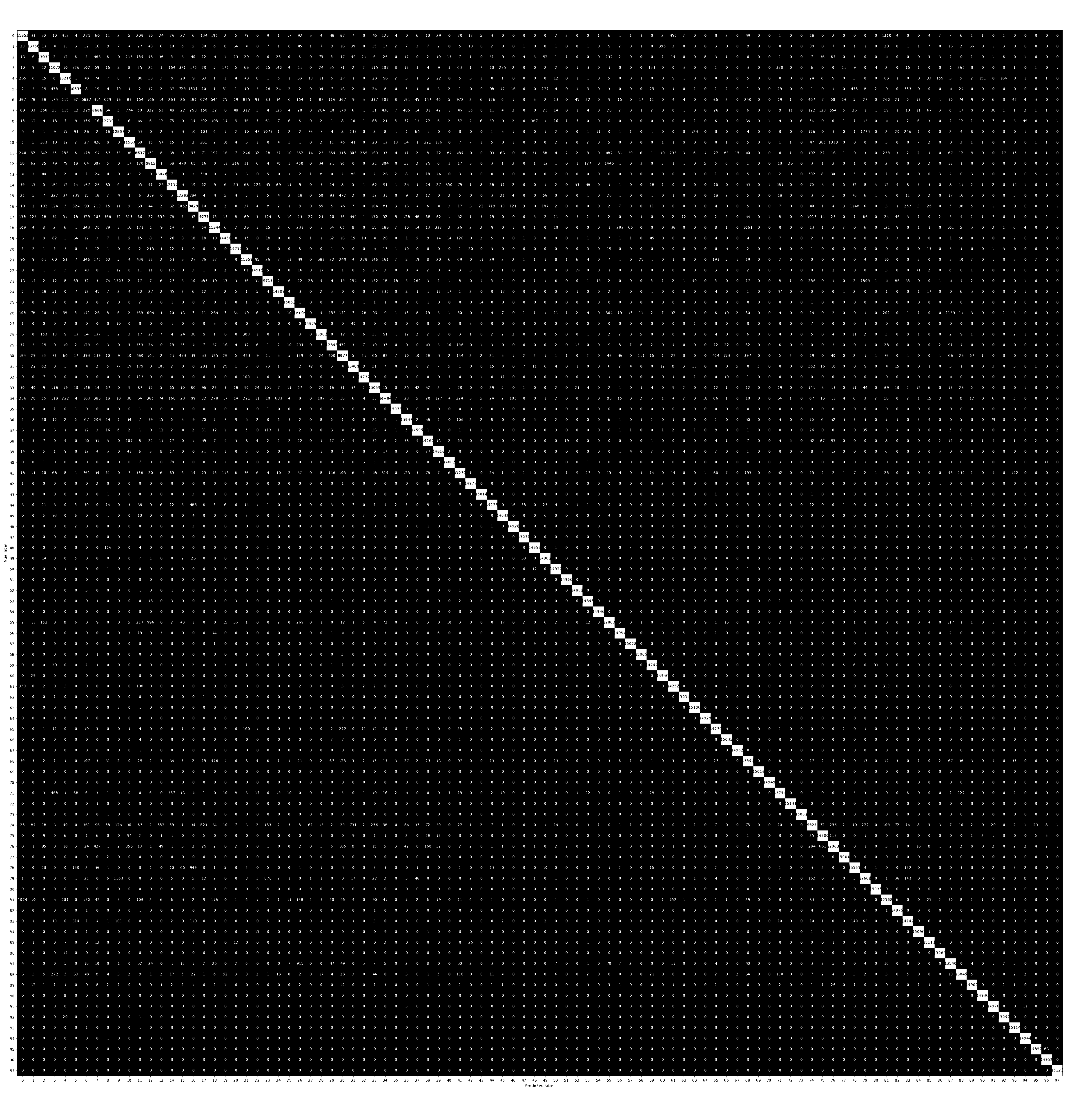}\\
  \caption{Confusion Matrix}
  \label{fig:confusion_matrix}
\end{figure}

The Confusion Matrix allowed for a detailed exploration of classification outcomes, delineating true positives, true negatives, false positives, and false negatives. This analysis unveiled a notable trend: the majority of errors observed stemmed from the disparities present in the real-world data's structure and semantics. This observation can be attributed to the inherent diversity and complexity of genuine medical texts, where nuances in language and context can lead to intricate classification challenges.

Interestingly, when the model was evaluated using synthetic data, the Confusion Matrix demonstrated a contrasting pattern. Synthetic data, crafted to adhere to specific structures and semantics, presented fewer challenges for the model's classification accuracy. This stark difference suggests that the model might encounter difficulties when confronted with the heterogeneity inherent in genuine medical text, compared to the more controlled environment of synthetic data.

The primary limitations observed encompass various aspects. First, the text length constraint, set at 128 words, significantly affects the model's ability to capture intricate nuances present in longer textual data. When exceeding this threshold, the data becomes represented as sparse vectors, potentially leading to information loss and diminished performance.

Secondly, the distinctive writing style encountered in the test data, which differs from that seen in the training data, poses a challenge. The model's training on a particular style limits its adaptability to new, previously unseen writing patterns. This mismatch between training and test data styles can result in reduced accuracy and nuanced misclassifications.

Moreover, the presence of questions addressing topics that were not covered extensively in the training data presents another constraint. Models struggle when faced with questions that delve into unfamiliar territories, as they lack the contextual familiarity to provide accurate predictions.

In conclusion, the discussed limitations, including text length constraints, writing style divergence, and unfamiliar thematic areas, collectively highlight the challenges faced when applying classifiers to such datasets. These limitations underscore the need for robust pre-training, data augmentation, and model fine-tuning strategies to enhance the model's performance and mitigate the observed shortcomings.

\section{Conclusion}
In this study, we collected a comprehensive dataset for text classification, augmented it with various techniques, and conducted experiments using five state-of-the-art transformer models: SBERT, BERT, LaBSE, BART, and RuBERT. We observed that RuBERT achieved the best performance with f1-score of 91.9\%, outperforming the other models. Based on these findings, we conclude that transformer models, particularly RuBERT, are highly effective for text classification tasks. The ability of transformers to capture contextual information and learn complex patterns in textual data contributes to their superior performance compared to classical machine learning methods.

Further research can be conducted to explore the performance of these transformer models on smaller datasets or specific domain-related datasets. Additionally, there is potential for developing new transformer architectures tailored specifically for text classification tasks. These architectures can incorporate domain-specific knowledge and enhance the model's ability to extract meaningful features from text, further improving classification accuracy.

Investigating transfer learning techniques, fine-tuning strategies, and hyperparameter optimization for these transformer models can also be valuable directions for future work. The exploration of different augmentation techniques and their impact on model performance can provide insights into improving the robustness and generalization capabilities of text classification models. Overall, there is ample opportunity for advancing the field of text classification using transformer models, and these future works can contribute to the development of more accurate and efficient models for various applications.

\bibliographystyle{splncs.bst}
\raggedright
\bibliography{bmf}
\end{document}